\documentclass[journal]{IEEEtran}

\ifCLASSINFOpdf
\else
   \usepackage[dvips]{graphicx}
\fi
\usepackage{url}

\usepackage{algorithmicx}
\usepackage{algpseudocode,algorithm}
\usepackage{caption}
\usepackage{graphicx}
\usepackage{microtype}
\usepackage{subfigure}
\usepackage{booktabs} % for professional tables
\usepackage{multirow}

\begin{document}

\title{A High-Performance Object Proposals based on Horizontal High Frequency Signal}

\author{Jiang Chao, Liang Huawei and Wang Zhiling
\thanks{``This work was supported by the Key Supported Project in the Thirteenth Five-year Plan of Hefei Institutes of Physical Science, Chinese Academy of Sciences under Grant KP-2017-35/KP-2017-13/KP-2019-16, the Independent Research Project of Research Institute of Robotics and Intelligent Manufacturing Innovation, Chinese Academy of Sciences under Grant C2018005,and the Youth Innovation Promotion Association of the Chinese Academy of Sciences under Grant 2017488.'' }
%\thanks{JiangChao is with the Hefei Institutes of Physical Science, Chinese Academy of Sciences, Anhui, China, and also with the University of Science and Technology, Anhui, China,(e-mail: jc2009@mail.ustc.edu.cn).}
\thanks{JiangChao,Liang Huawei and Wang Zhiling are with Hefei Institutes of Physical Science, Chinese Academy of Sciences, the University of Science and Technology of China, the Anhui Engineering Laboratory for Intelligent Driving Technology and Application, and also with the Innovation Research Institute of Robotics and Intelligent Manufacturing, Chinese Academy of Sciences, Anhui, China,(e-mail: hwliang@iim.ac.cn).} 
\thanks{(Corresponding author: Liang Huawei and Wang Zhiling.)}}

\markboth{Journal of \LaTeX\ Class Files, Vol. 16, No. 9, August 2020}
{Shell \MakeLowercase{\textit{et al.}}: Bare Demo of IEEEtran.cls for IEEE Journals}
\maketitle

\begin{abstract}
In recent years, the use of object proposal as a preprocessing step for target detection to improve computational efficiency has become an effective method. Good object proposal methods should have high object detection recall rate and low computational cost, as well as good localization quality and repeatability. However, it is difficult for current advanced algorithms to achieve a good balance in the above performance. For this problem, we propose a class-independent object proposal algorithm BIHL. It combines the advantages of window scoring and superpixel merging, which not only improves the localization quality but also speeds up the computational efficiency. The experimental results on the VOC2007 data set show that when the IOU is 0.5 and 10,000 budget proposals, our method can achieve the highest detection recall and an mean average best overlap of 79.5\textsc{$\%$}, and the computational efficiency is nearly three times faster than the current fastest method. Moreover, our method is the method with the highest average repeatability among the methods that achieve good repeatability to various disturbances.

\end{abstract}


\IEEEpeerreviewmaketitle

\section{Introduction}

\label{submission}

\begin{figure}[h]
\vskip -0.1in
\centering
{
\begin{minipage}[t]{0.266\linewidth}
\centering
\includegraphics[width=\columnwidth]{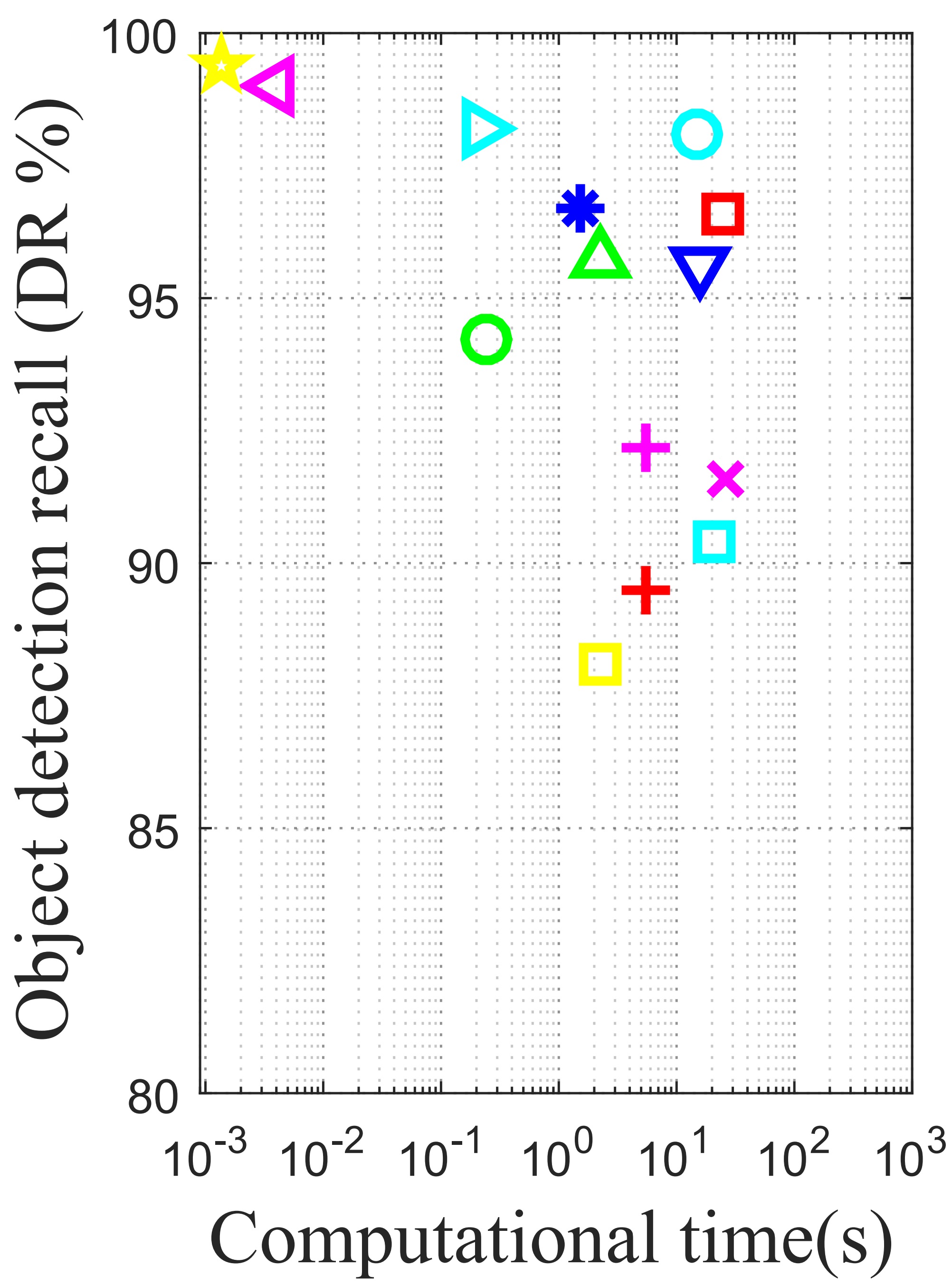}
%\caption{fig1}
\end{minipage}%
}%
{
\begin{minipage}[t]{0.252\linewidth}
\centering
\includegraphics[width=\columnwidth]{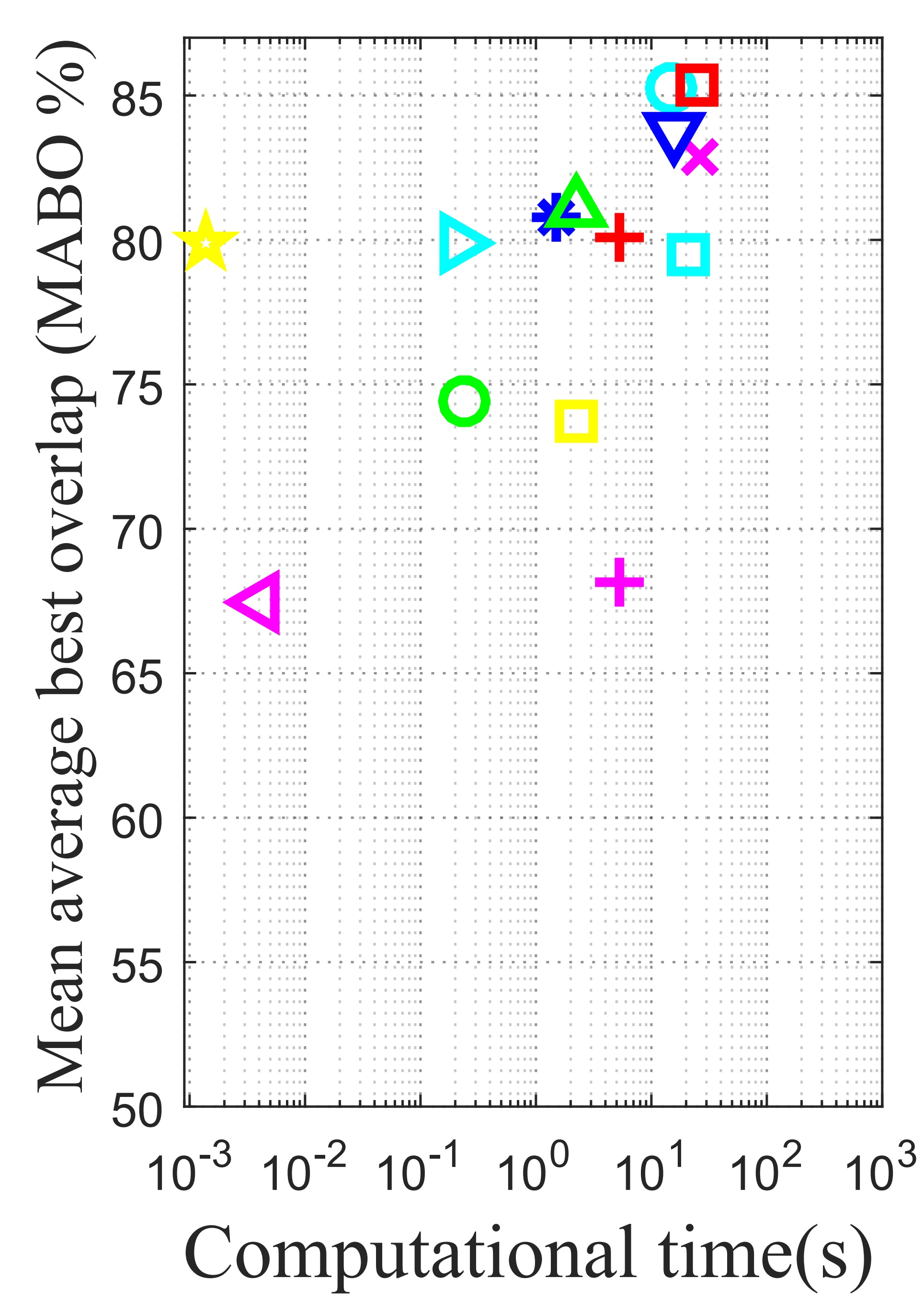}
%\caption{fig2}
\end{minipage}%
}%
{
\begin{minipage}[t]{0.246\linewidth}
\centering
\includegraphics[width=\columnwidth]{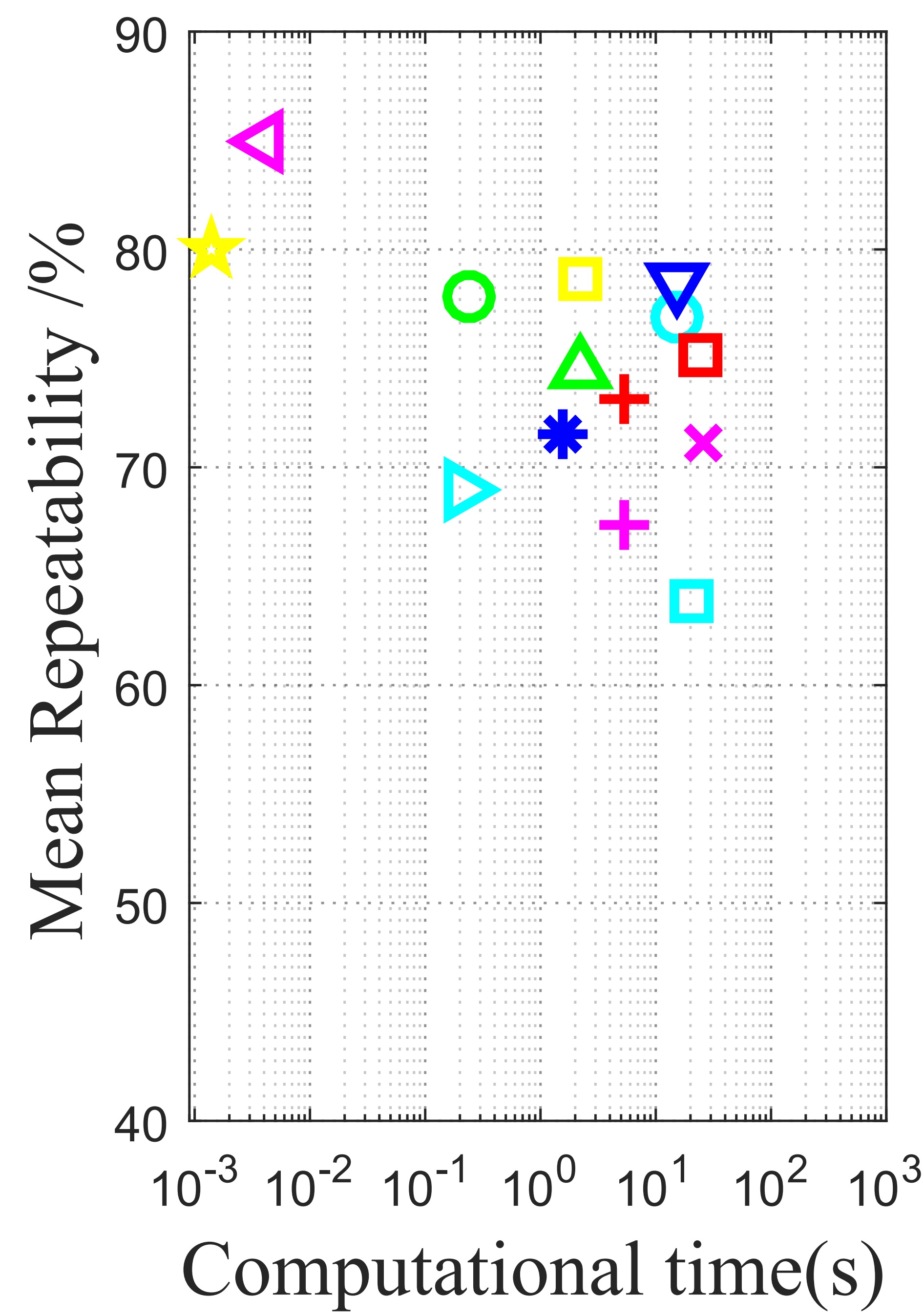}
%\caption{fig2}
\end{minipage}
}%
{
\begin{minipage}[t]{0.12\linewidth}
\centering
\includegraphics[width=0.36in]{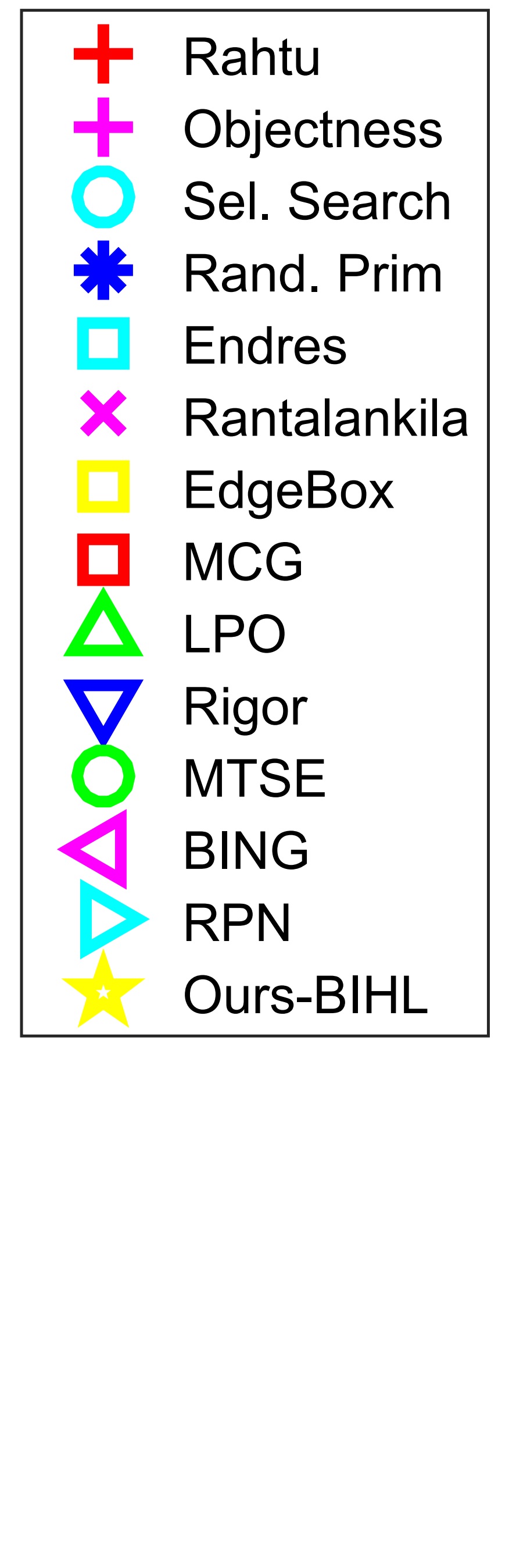}
%\caption{fig2}
\end{minipage}
}%
\centering
\begin{center}
\caption{Comparison of the most advanced general object proposal methods on the VOC2007 data set [1] when the proposal is $10^4$ and the overlap threshold $IOU\geq0.5$. Our approach achieves the best performance between recall (DR), repeatability and computational efficiency, and MABO is similar to that of the most advanced methods. All competition results are generated by public code (see the details in our experiment section).}
\end{center}
\vskip -0.3in
\end{figure}

\IEEEPARstart{U}{sing} sliding Windows to extract densely overlapping detection boxes and then analyze them is a popular strategy in object detection methods [2]. However, in general object detection tasks, in order to obtain good detection accuracy, a large number of potential windows need to be detected. For example, images of $W*H$ pixels, the number of potential windows can reach $O(W^2H^2)$, so that most of the computing resources are wasted on useless potential windows. In recent years, a variety of general object proposal schemes have been proposed and attracted extensive attention, based on the visual attention mechanism of the human eye. The general object proposal can be accurately generated under the premise that the type of the detected object is uncertain, which can reduce the number of search windows in the preprocessing stage of object detection and recognition, thereby greatly reducing the computational time. This scheme has been adopted by mask R-CNN [3], faster R-CNN [4], Cascade R-CNN [5] and other excellent target detection algorithms [2]. The results show that the object proposal scheme can be successfully applied to complex computer vision systems as a data preprocessing step. Among them, Sel.search [6], Edgeboxes [7] and BING [8] have the most advanced performance.Object proposal as a pre-processing step for other algorithms must meet the highest possible object detection recall rate in the shortest possible time to avoid missed detection and improve real-time performance. Among them, only BING[8] can achieve optimal performance in terms of recall rate and computational efficiency, which is why BING is widely used in various image detection [9].

However, the localization quality of the proposals generated by BING is not high, and the recall rate also drops rapidly when the overlap threshold IOU is stricter.  Although Sel.search[6], MCG [12] and other super-pixel merging methods can achieve better performance in localization quality, but at the cost of great calculation time. The problems of these two types of methods can be summarized as low efficiency and poor localizatio quality. For this problem, we have proposed a solution that can significantly improve the localization quality and computational efficiency, while also improving the recall rate and the robustness to noise interference. Specifically, our main work and contributions:
\begin{itemize}
\item We propose to use horizontal high frequency features (HL) to describe closed contours: HL features have better objectness measuring and lower computational complexity than NG features, which makes our algorithms have higher recall rates and higher computational efficiency. The test results in Pascal VOC2007 [1] show that the calculation of HL feature is only one eighth of NG feature, while the recall rate of HL feature is higher than NG feature.
\item We propose a efficient eight-neighbor edge tracking and bounding box growth algorithm to merge window borders containing different parts of the same object: by iteratively merging neighborhood boxes with the same attributes, the merged proposal box is closer to the real area. On the VOC2007 data set, using this strategy we can make the BIHL algorithm improve the localization quality by 8.7$\%$ without reducing the recall rate, and the calculation time is only 0.04ms.
\end{itemize}

\section{Proposed Approach}
In this section, we will describe our method of generating an object proposal. The implementation process of the algorithm is shown in Figure 3.\\
First, for each test image I, downsampling is performed according to a series of fixed ratios  $P_{ds}$, where $(m, n)$ respectively represent the ratio of image line and column downsampling.
\vskip -0.02in
\begin{equation}
P_{ds}\left (m,n \right )=[2^m,2^n ], \left\{
             \begin{array}{l}
             if \; m \in [0,1,2,3],|n-m| \leq 2   \\

             if \; m=4,n=3
             \end{array}
\right.
\end{equation}
\vskip -0.02in

Secondly, for each downsampled image $I_{P_{ds (m,n)}}$, we calculate its horizontal high-frequency HL feature map $I_{F_{ds (m,n)}}$, and traverse each feature map $I_{F_{ds (m,n)}}$ with a fixed-size search template. We limit the maximum downsampling ratio during downsampling, which reduces the number of times the search template traverses the image.Since each downsampled image needs to be traversed using a search template, the less the number of downsamplings, the smaller the amount of computation. Although the area of the window we initially generated cannot contain large objects, we don't need to worry about this, because we have adopted a border merging strategy for these proposals, and generated a high-quality window that approximates the real area of the object by tracking the edge of the object. These new proposals include large-scale objects.The border merging strategy is an iterative fusion process.The main purpose of this strategy is to improve the poor quality of the window scoring method, which is an insurmountable defect. Since our border merging process only processes a small number of window borders with the same attributes, the amount of calculation is almost negligible compared to other steps of the algorithm, which we will introduce later.Third, for each region of the search template traversal, we use a binarized classifier to score it, and perform non-maximum suppression and border merging based on the scored results.\\

\begin{figure}[ht]
\vskip -0.1in
\begin{center}
%\centerline{\includegraphics[width=0.9\columnwidth]{flowchart.pdf}}
\includegraphics[width=0.7\columnwidth]{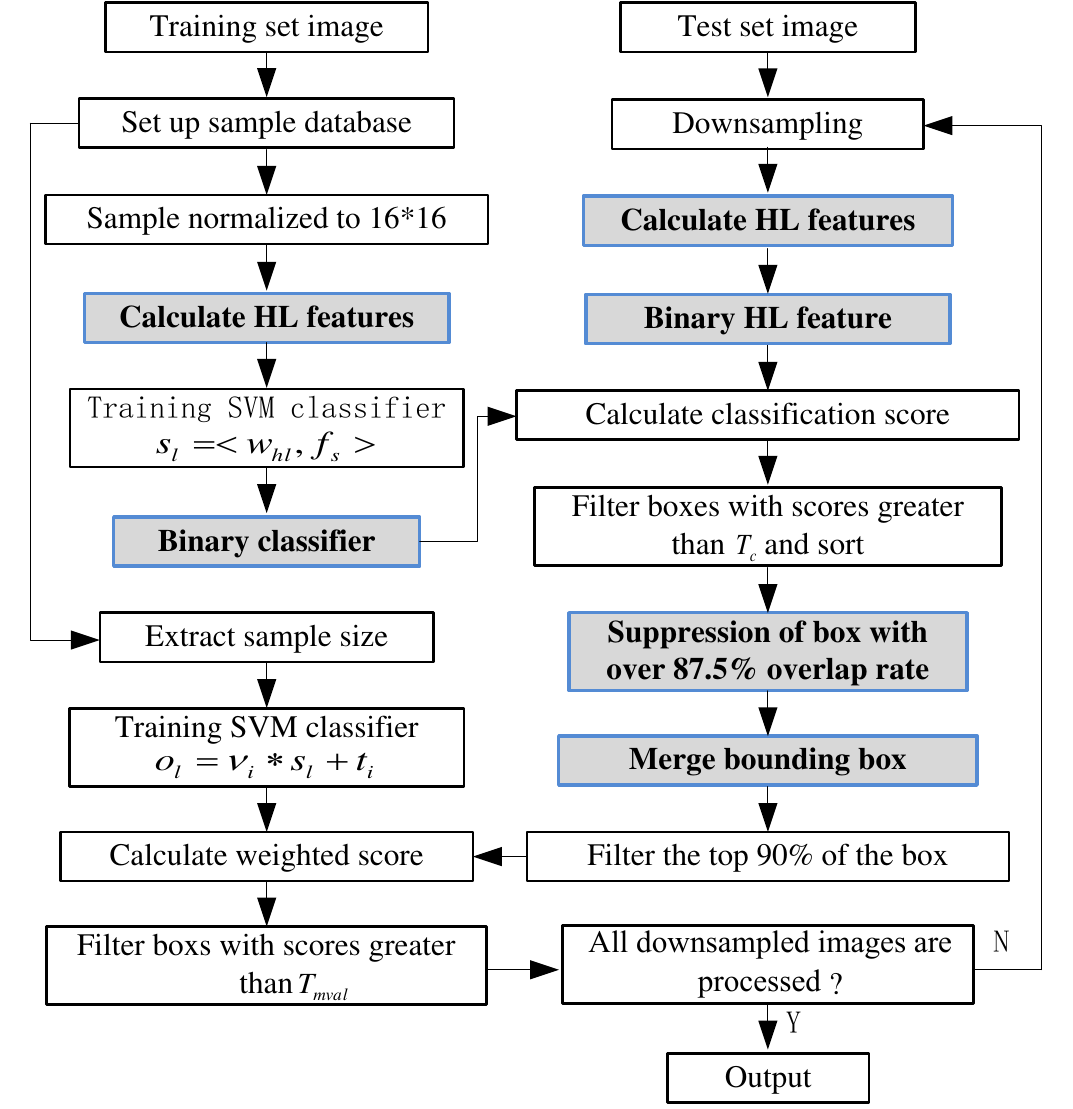}

\caption{Implementation flow chart of BIHL algorithm.}
\end{center}
\vskip -0.2in
\end{figure}

In Figure 2, the thresholds $ T_ {c} $ and $ T_ {mval} $ are set to filter out background interference.

\subsection{HL frequency feature and Object proposal}
Objects are usually rich and complex in color and texture, so the method of using color and gradient features such as BING has a higher recall rate at a lower overlap threshold, but the localization quality is lower. And as the overlap threshold increases, its recall rate will drop dramatically. Frequency signals has good ability to extract image structure information, and has a wide range of applications in image processing field. We have found that this feature of frequency can be applied to the feature extraction of the object proposal generation method to achieve good results.\\
Calculation method: for the down sampled image $I_{P_{ds(m,n)}}$, first use the high pass filter $f_{h}$ defined in equation 2 to convolute the image in the row direction in steps of 2, and then use the low pass filter $f_{l}$ defined in equation 3 to deconvolute the image in the column direction in the same steps, then the horizontal high frequency feature map $I_{F_{ds (m,n)}}$ can be obtained.
%$$f_{h}=[\ 1 \ \ \; 1]/1.41 \eqno{(1)}$$
%$$f_{l}=[-1 \; 1]/1.41 \eqno{(1)}$$
\begin{equation}
  \begin{array}{l}
   f_{h}=[-1 \; 1]/1.41 
  \end{array}
\end{equation}
\begin{equation}
  \begin{array}{l}
   f_{l}=[\ 1 \ \ \; 1]/1.41 

  \end{array}
\end{equation}
\vskip -0.01in
As shown in Table I, the HL feature requires only a few basic operations to achieve. We counted all the required basic operands and compared it with BING's NG feature.
\begin{table}[htbp]
\vskip -0.0in
\begin{small}
 \centering

\setlength{\belowcaptionskip}{-0pt}\centering\caption{\label{tab:test}The basic operands of HL feature in the image of w * h.}
\setlength{\tabcolsep}{0.36mm}{
 \begin{tabular}{lcccccccr}

  \hline

% \tabincell{c}{Variants} & \tabincell{c}{Calculating \\ feature /ms} & \tabincell{c}{Classifica- \\tion /ms} & \tabincell{c}{ Suppression \\ \& merging /ms} & \tabincell{c}{Others \\ /ms} & \tabincell{c}{Total \\ /ms} \\

  \hline

Feature & + & - & * & / & abs & max & min \\
HL & 5wh & 1.5wh & 2wh & 1.5wh & 0 & 0 & 0 \\
NG & 7wh & 20wh+5w+h & 4wh+4h+4w & 4wh-4w-4h & 14wh & 4wh & 3wh \\

  \hline

 \end{tabular}}
\end{small}
\vskip -0.01in
\end{table}
\\
The basic operands of HL feature are only 1 / 8 of NG feature, so theoretically, the computing efficiency of HL feature is much higher than NG feature.\\
We use the HL as the metric feature of objectness. Calculate the horizontal high-frequency HL feature map $I_{F_{ds(m,n)}}$ on a series of downsampled images, and then a sliding window is used to traverse each $8*8$ scale region on the feature map in steps of 1 to obtain the 64 dimensional HL feature $hl$ of the corresponding region. Then we binarize each $hl$ feature and use the binary SVM classifier to calculate the classification score.

\subsection{Binarized}
\subsubsection{HL feature binarization}
The binarized expression of the HL feature $f_{s}$ is as shown in Eq.$(4)$, and the first $N_{g}$ binary bits are used to approximate the 64-dimensional HL frequency feature $hl$ [8], Where $k$ represents a binary bit.
\begin{equation}
  \begin{array}{l}
 f_{s}=\sum_{k=1}^{N_{g}} 2^{8-k}hl_{k}
  \end{array}
\end{equation}
\subsubsection{Classifier binarization}

Literature[8][20] explains that binary approximation of classifier has the advantage of accelerating feature extraction and classification process. In order to improve the computational efficiency, we refer to [8] [20] using a set of basic vectors to approximate the SVM classifier model $w_{hl}$.
%\vskip -0.1in
%\begin{algorithm}[hbt]
%\begin{scriptsize}
%   \caption{Binary approximation of $w_{hl}$}
%   \label{alg:example}
%\begin{algorithmic}
%   \State {\bfseries Input:} $w_{hl}$,$N_{a}$,$N_{b}$,$T_{d}$
%   \State {\bfseries Output:} $\{\lambda_{i}\}_{i=1}^{N_{a}}$,$\{\nu_{i}\}_{i=1}^{N_{a}}$,$\{s_{i}\}_{i=1}^{N_{b}}$,$\{\mu_{i}\}_{i=1}^{N_{b}}$
%   \State $\delta=w_{hl}$ (initialize residual).
%   \For{$i=1$ {\bfseries to} $N_{a}$}
%   \State $\nu_{i}=Sign(\delta)$
%   \State $\lambda_{i} = \langle{\nu_{i} , \delta}\rangle/||\nu_{i}||^2$
%   \State $\delta \leftarrow \delta-\lambda_{i}\nu_{i} $ (update residual)
%   \EndFor
%   \State $\rho=find(abs(\delta)>T_{d})$ .
%   \State $N_{b}=length(\rho)$ .
%   \For{$i=1$ {\bfseries to} $N_{b}$}
%   \State $s_{i}(\rho_{i})=-Sign(\delta(\rho_{i}))$
%   \State $\mu_{i}(\rho_{i}) = abs(\delta(\rho_{i}))$
%   \EndFor
%\end{algorithmic}
%\end{scriptsize}
%\end{algorithm}
\vskip -0.1in
\begin{equation}
  \begin{array}{l}
w_{hl} \approx \sum_{i=1}^{N_{a}} \lambda_{i} \nu_{i}
  \end{array}
\end{equation}
Using this approximation, we can effectively compute the dot product $\langle{w_{hl} , f_{s}}\rangle$ using only bitwise operations.\\
Where: $\nu_{i}$ can be represented by a binary vector, $\nu_{i}=2\nu_{i}^{+}-1$ , $\nu_{i}^{+} \in \{0,1 \}^{D}$ .The following ex-pression can be obtained:\\
\begin{small}
\begin{equation}
  \begin{array}{l}
\! \langle{w_{hl} , f_{s}}\rangle \approx \! 
 \sum\limits_{i=1}^{N_{a}} \lambda_{i} \sum\limits_{k=1}^{N_{g}} 2^{8-k} (2 \langle{\nu_{i}^{+} , h_{k}}\rangle -|h_{k}|)
  \end{array}
\end{equation}
\end{small}
\vskip -0.1in
Where $\lambda_{i} \in R$ represent the calibration coefficients, $N_{a}$  represents the number of decomposition basis vectors. 

\subsection{Non-maximum suppression and Boxes Merging }
Using the Eq.(6) to traverse the feature map $F_{P_{ds(m,n)}}$, a classifier score matrix M can be generated, and each element value of the matrix M represents the objectness score of the $16*[2^m,2^n ]$ region of the corresponding original image, where $(m, n)$ respectively represent the ratio of image line and column downsampling. In this paper, the non-maximum suppression algorithm is used to filter the background region where the score is less than 0 in $M$ and suppress the window with the overlap ratio greater than 87.5$\%$. The proposal box coordinates and scores after the non-maximum suppression are stored in the container $V$, and the merge strategy proposed by us is used to merge the borders.

\subsubsection{Bounding box merge }

The proposal generated by a fixed-size search template on a series of downsampled images is likely to not well surround the object instance, resulting in a lower quality of the final generated proposal boxes.In order to solve this problem, we propose a fast and effective eight-neighborhood edge tracking and border growth algorithm to merge box borders containing different parts of the same object. Our method only fuses a small number of box borders with the same attributes, which takes almost no time. We merge the bounding boxes after non-maximum suppression. The merging process is based on the principle that the scores and positions are similar.
The steps to implement the algorithm are as follows:
\begin{itemize}
\item Create an all-one matrix $M_1$. The proposals stored in the container V are projected into $M_1$ according to the score from high to low, and the matrix value of the projected coordinates in $M_1$ becomes 0, and the projection point is used as a seed point $A$ to be grown;

\item In $M_1$, judge whether the value of 8 fields around the current seed point $A$ is 0, if not, set 0, if 0, it means that the point has been grown by the previous seed point $B$, then judge whether the difference between the serial number of $A$ in $V$ and the serial number of $B$ in $V$ is less than the threshold value $T_{s1}$, if less than, fuse the two seed points, otherwise as a new seed point;

\item If there is seed point $B$ on the coordinate point when proposal $C$ is projected to $M_1$, judge whether the difference between the serial number of $C$ in $V$ and the serial number of $B$ in $V$ is less than the threshold value $T_{s2}$. If it is less than the threshold value, fuse the two seed points, otherwise delete proposal $C$;
\item We limit the maximum length of $V$ to 1100, and stop growing when there is no proposal in the container $V$.
\end{itemize}

%\begin{figure}[ht]
%\vskip -0.1in
%\begin{center}
%\centerline{\includegraphics[width=0.7\columnwidth]{bordergrowthimage.jpg}}
%\caption{Schematic diagram of eight neighborhood edge tracking and border growth algorithm.}
%\end{center}
%\vskip -0.3in
%\end{figure}

%Where $T_{s1}$ and $T_{s2}$  are the distance thresholds of the objectness score.
\subsubsection{Performance evaluation }
In order to verify the performance of the merge strategy, we tested the recall rate (DR), localization quality (MABO) and computation time (Time) when using the merge strategy and disabling the merge strategy on the VOC2007 dataset [1].

\newcommand{\tabincell}[2]{\begin{tabular}{@{}#1@{}}#2\end{tabular}}

\begin{table}[htbp]
\vskip -0.0in
\begin{small}
 \centering

\caption{\label{tab:test}When the number of proposals is 10000 and the overlap threshold is 0.5, compare the performance when the merge strategy is adopted and disabled..}

\setlength{\tabcolsep}{1.2mm}{
 \begin{tabular}{lccc|cccr}

  \hline

% \tabincell{c}{Variants} & \tabincell{c}{Calculating \\ feature /ms} & \tabincell{c}{Classifica- \\tion /ms} & \tabincell{c}{ Suppression \\ \& merging /ms} & \tabincell{c}{Others \\ /ms} & \tabincell{c}{Total \\ /ms} \\

  \hline
Methods	& \multicolumn{3}{l}{No merge strategy} & \multicolumn{3}{l}{Use merge strategy} \\
\hline
Indicators & DR & MABO & Time & DR & MABO & Time  \\

 \hline
BIHL  & 99.2$\%$ & 70.8$\%$ & 1.49ms & 99.2$\%$ & 79.5$\%$ & 1.54ms \\
\hline
BING  & 99.1$\%$ & 69.9$\%$ & 4.0ms & 99.1$\%$ & 79.2$\%$ & 4.41ms\\

  \hline

 \end{tabular}}
\end{small}
\vskip -0.1in
\end{table}

According to Table II, the recall rate does not change when the BIHL algorithm adopts the merge strategy, while the localization quality has greatly improved, and the merge strategy only accounts for about 3$\%$ of the calculation time (0.05ms). When BING  adopted our merger strategy, the localization quality improved by 9.3$\%$.

\begin{table*}[htp]
\vskip -0.01in
\begin{scriptsize}
 \centering

\caption{\label{tab:test}MABO and Object Detection Rate Comparison on VOC2007 test set.}
\setlength{\tabcolsep}{2.12mm}{
 \begin{tabular}{l|cccccccccccccccc|ccr}
   \hline
\multicolumn{2}{l}{Methods}	& EB5 & EB9 & E & LOP & ME & MCG & Obj & R.P & U & Ru & R1 & $Sel.S$ & Rr & BING & RPN& \tabincell{c}{BIHL}  \\
 \hline

\multicolumn{2}{l} {$\# prop$} & 1091 & 9999	 & 1620 & 1114 & 1410	& 1939	& 1772	& 5745	& 9999 	&  9999	& 7689	& 8753	& 1536	& 9818 & 2000  & 9797	 \\
\multicolumn{2}{l}{MABO/$\%$} & 73.7 & 82.4 & 79.5	& 81.2	& 74.4	& \textbf{\underline{85.4}}& 68.2	& 80.8	& 79.0  & 80.1 & 82.9 & \textbf{\underline{85.3}}	& \textbf{\underline{83.8}}	& 69.9 & 79.9  & 79.5  \\

\multicolumn{2}{l} { \tabincell{c}{Repeatability}} & \textbf{\underline{0.79}} & 0.75  & 0.64 & 0.75 & 0.78	& 0.75	& 0.67	& 0.72	& 0.65  & 0.73	& 0.71 & 0.77 & 0.78 & \textbf{\underline{0.85}} & 0.69  & \textbf{\underline{0.8}}  \\
  
\multicolumn{2}{l} { \tabincell{c}{DR/$\%$}}  & 96.7 & 88.2  & 90.4 & 95.7 & 94.2 & 96.7 & 92.2 & 96.7 & 92.0  & 89.5 & 91.6 & 98.1 & 95.6 & \textbf{\underline{99.1}}  & \textbf{\underline{98.2}} & \textbf{\underline{99.2}} \\
 \multicolumn{2}{l} { \tabincell{c}{Time/s}} & 2.23 & 2.24  & 150.3 & 2.25 & 0.24 	& 24.9	& 5.4	& 1.54	& /  & 5.47	& 26.15 & 15.13 & 15.73 & \textbf{\underline{0.004}} & \textbf{\underline{0.22}}  & \textbf{\underline{0.0015}}  \\
  \hline
 \end{tabular}}
\end{scriptsize}
\vskip -0.01in
\end{table*}

\section{Experiment}

To verify the validity and superiority of the proposed method, we tested our method in the public dataset Pascal VOC2007 [1] and the Pascal VOC2007 synthetic interference dataset [2] with 297,120 test images [2],and compare it with the best typical method in the last decade:MCG[12]$\footnote{ https://github.com/batra-mlp-lab/object-proposals. \label{1} }$ , Sel.Search(Sel.S)[6]\textsuperscript{\ref{1}}, Rand.Prim(R.P)[13]\textsuperscript{\ref{1}}, MTSE (ME)[17]$\footnote{ http://3dimage.ee.tsinghua.edu.cn/cxz/mtse. \label{2} }$, Endres(E)[11]\textsuperscript{\ref{2}}, LOP[14]$\footnote{ https://github.com/hosang/detection-proposals. \label{3} }$, Rantalankila (R1)[16]\textsuperscript{\ref{3}}, Rigor (Rr)[15]\textsuperscript{\ref{3}}, Objectness(Obj)[18]\textsuperscript{\ref{3}}, BING[8]\textsuperscript{\ref{3}}, EdgeBox (EB)[7]\textsuperscript{\ref{3}}, Rahtu (Ru)[19]\textsuperscript{\ref{3}} , Uniform(U)[2]\textsuperscript{\ref{3}},RPN[4]$\footnote{ https://github.com/endernewton/tf-faster-rcnn. \label{4} }$.VOC2007 contains 20 object categories, including 9,963 natural images with annotated labels. We trained using a training data set consisting of 2501 images and tested on a test data set consisting of 4952 images.The VOC2007 Interference Dataset [2] is a repeatability test set synthesized by adding scaling, rotation, illumination variation, JPEG artefacts, blurring, and saltnpepper noise in the VOC2007 test set[1].
The four key evaluation indicators of the general object proposal methods are recall rate , computational efficiency, localization quality, and repeatability[2]. The recall rate evaluates whether it is possible to use as few object proposals as possible including as many objects as possible in the image. Computational efficiency evaluates whether the speed of the object proposal method is significantly faster than the detector itself, so as to improve the computational efficiency of the detector. Repeatability evaluates whether the performance of the object proposal method remains stable on similar or slightly different images. Localization quality evaluates the validity of the generated object proposal. Therefore, we also mainly test and compare four aspects: (1) object detection recall (DR), (2) mean average best overlap (MABO), (3) computing time, and (4) repeatability .\\
Because these methods have been optimized for the Pascal VOC2007 dataset[1], we only use their default parameter settings. It is worth noting that the VOC2007 data set is labeled with simple samples and difficult samples. In this paper, the comparison is made under the condition of selecting all difficult samples and simple samples.

\subsection{Recall}
When the detection and recognition algorithm uses box generated by the object proposal method as a candidate window, whether the object proposal method can contain all the interested objects in the test image is the most critical criterion to measure their performance.We evaluate by calculating the maximum recall rate that can be achieved under a fixed overlapping thresholds (IoU) and the number of proposals is less than 10,000 [2].In the experiment we used the overlapping thresholds $IOU\geq 0.5 $ to calculate the recall rate. The experimental results are shown in Table 3.

\subsection{MABO}
In order to evaluate the quality of our object proposals, Mean average best overlap (MABO) is used to measure the ability of how well the proposals generated by a method can localize object instances [2].

In Table III, we summarize the recalls and MABOs for each method at different overlap thresholds when the number of proposals is less than , and underline the top 3 methods in performance. Our method, BIHL, achieves the highest recall rate in all methods at $IOU\geq 0.5 $. The MABO of our method is slightly lower than that of some segmentation method and superpixel merging method. But when the localization quality reaches 80$\%$, the performance of the proposed algorithm depends more on the computational efficiency, detection recall rate and repeatability to perturbation.

\subsection{Proposal Repeatability}
By referring to the evaluation method of Jan Hosang et al. [2], we compared BIHL with other most advanced algorithms, and the results are shown in figure 3. 

In Table III we summarize the average repeatability of all state-of-the-art methods.The repeatability of various perturbation is generally poor with RPN[4] and segmentation methods and superpixel merging methods. The average repeatability of BIHL is second only to BING[8]. However, as shown in Figure 3, BING are very sensitive to scaling, but BIHL can still achieve good performance.This shows that BIHL is the method with the highest average repeatability among the state-of-the-art methods that maintain good repeatability to various perturbation.\\
\begin{figure}[ht]
\vskip -0.1in
\begin{center}
%\centerline{\includegraphics[width=0.9\columnwidth]{flowchart.pdf}}
\includegraphics[width=0.9\columnwidth]{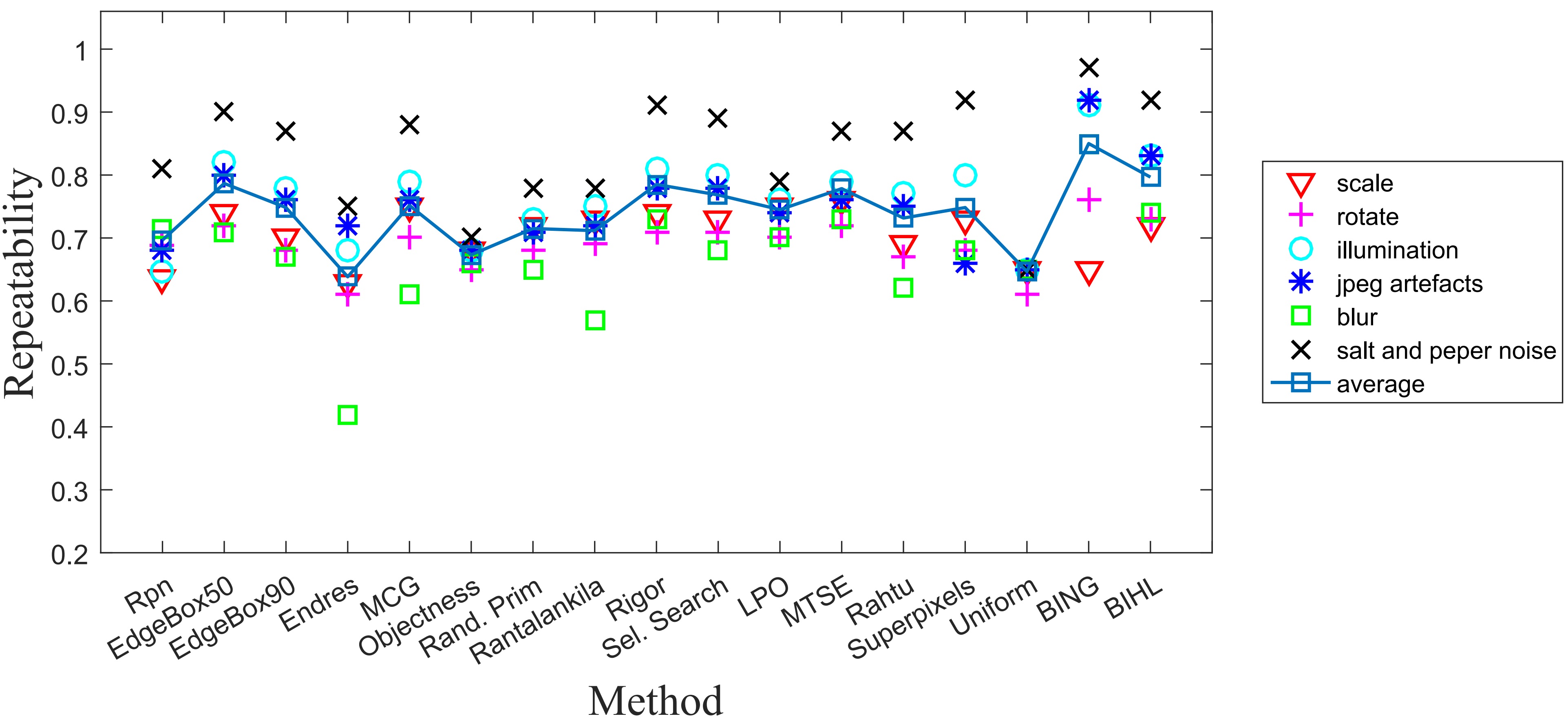}

\caption{Repeatability results under different perturbations.}
\end{center}
\vskip -0.3in
\end{figure}

\subsection{Computational Time}
Table III shows the computational time of all advanced methods on the PASCAL VOC2007 test set containing 4952 images [1]. RPN[4] runs on GTX 1080Ti GPU and 2.20 GHz Intel Xeon E5-2630 v4 CPU. Other methods only run on Intel Xeon E5-2630 v4 CPU. The time shown here is the average time of all images of the PASCAL VOC2007 test set[1], including all steps from reading the image to outputting the proposals.In general, segmentation-based methods and superpixel merging methods run relatively slowly. For example, the fastest method Rand. Prim[13] also requires 1.54 s, while the slowest method Endres[11] requires 150s. Edge-based methods are usually less computationally intensive, For example, the slowest method Rahtu[19] only requires 5.47 s, and the fastest BING[8] of them can reach 0.004 s. The three variants of our method, except for the difference in accuracy and proposal budget, run at almost the same speed, 0.0015s per image, which is the most efficient method of all current methods, which is nearly 3 times faster than BING [8].

\section{Conclusion}
In this paper, we propose a new object proposal algorithm(BIHL), which can generate high quality proposals in a very effective way. The advantages of the algorithm are mainly brought by two parts: (1) HL frequency feature, (2) bounding box merging. BIHL has a minimal computational complexity and good objectness measurement ability,  which make our algorithm achieve the best performance of the recall rate and computational efficiency in all object proposal algorithms on the VOC2007 dataset, and the average optimal overlap rate (MABO) reaches 79.5$\%$. At the same time, BIHL is the method with the highest average repeatability among the state-of-the-art methods that maintain good repeatability to various perturbation. The code will be published in \url{https://github.com/JiangChao2009/BIHL}.

%\section*{References}


\medskip

\small
\begin{thebibliography}{20}
\bibitem{}Everingham, M., Van Gool, L., Williams, C. K., Winn, J., and Zisserman, A. The pascal visual object classes challenge 2007 (voc2007) results. 2007.

\bibitem{}Hosang, J., Benenson, R., Doll’ar, P., and Schiele, B. What makes for effective detection proposals? IEEE transactions on pattern analysis and machine intelligence, 38(4): 814-830, 2015.

\bibitem{}He, K., Gkioxari, G., Doll’ar, P., and Girshick, R. Mask rcnn. In Proceedings of the IEEE international conference on computer vision, pp. 2961-2969, 2017.

\bibitem{}Ren, S., He, K., Girshick, R., and Sun, J. Faster r-cnn: Towards realtime object detection with region proposal networks. In Advances in neural information processing systems, pp. 91-99, 2015.

\bibitem{}Cai, Z. and Vasconcelos, N. Cascade r-cnn: Delving into high quality object detection. In Proceedings of the IEEE conference on computer vision and pattern recognition, pp. 6154-6162, 2018. 

\bibitem{}Uijlings, J. R., Van De Sande, K. E., Gevers, T., and Smeulders, A. W. Selective search for object recognition. International journal of computer vision, 104(2):154-171, 2013.

\bibitem{}Zitnick, C. L. and Doll’ar, P. Edge boxes: Locating object proposals from edges. In European conference on computer vision, pp. 391-405. Springer, 2014.

\bibitem{}Cheng, M.-M., Zhang, Z., Lin, W.-Y., and Torr, P. Bing: Binarized normed gradients for objectness estimation at 300fps. In Proceedings of the IEEE conference on computer vision and pattern recognition, pp. 3286-3293, 2014.

\bibitem{}Yao, Y., Liu, P., Sun, X., and Zhang, L. Moving object surveillance using object proposals and background prior prediction. Journal of Visual Communication and Image Representation, 61:85-92, 2019.

\bibitem{}Felzenszwalb, P. F., Girshick, R. B., McAllester, D., and Ramanan, D. Object detection with discriminatively trained part-based models. IEEE transactions on pattern analysis and machine intelligence, 32(9):1627-1645, 2009.

\bibitem{}Endres, I. and Hoiem, D. Category-independent object proposals with diverse ranking. IEEE transactions on pattern analysis and machine intelligence, 36(2):222-234, 2013.


\bibitem{}Pont-Tuset, J., Arbelaez, P., Barron, J. T., Marques, F., and Malik, J. Multiscale combinatorial grouping for image segmentation and object proposal generation. IEEE trans-actions on pattern analysis and machine intelligence, 39 (1):128-140, 2016.

\bibitem{}Manen, S., Guillaumin, M., and Van Gool, L. Prime object proposals with randomized prim’s algorithm. In Proceed-ings of the IEEE international conference on computer vision, pp. 2536-2543, 2013.


\bibitem{}Krähenbühl, P. and Koltun, V. Learning to propose objects.In Proceedings of the IEEE Conference on Computer Vision and Pattern Recognition, pp. 1574-1582, 2015.

\bibitem{}Humayun, A., Li, F., and Rehg, J. M. Rigor: Reusing inference in graph cuts for generating object regions. In Proceedings of the IEEE Conference on Computer Vision and Pattern Recognition, pp. 336-343, 2014.

\bibitem{}Rantalankila, P., Kannala, J., and Rahtu, E. Generating object segmentation proposals using global and local search. In Proceedings of the IEEE conference on computer vision and pattern recognition, pp. 2417-2424, 2014.

\bibitem{}Chen, X., Ma, H., Wang, X., and Zhao, Z. Improving object proposals with multi-thresholding straddling expansion. In Proceedings of the IEEE conference on computer vision and pattern recognition, pp. 2587-2595, 2015.

\bibitem{}Alexe, B., Deselaers, T., and Ferrari, V. Measuring the objectness of image windows. IEEE transactions on pattern analysis and machine intelligence, 34(11):2189-2202, 2012.

\bibitem{}Rahtu, E., Kannala, J., and Blaschko, M. Learning a category independent object detection cascade. In 2011 international conference on Computer Vision, pp. 1052-1059. IEEE, 2011.

\bibitem{}Hare, S., Saffari, A., and Torr, P. H. Efficient online structured output learning for keypoint-based object tracking. In 2012 IEEE Conference on Computer Vision and Pattern Recognition, pp. 1894-1901. IEEE, 2012.
\end{thebibliography}
\end{document}